# 一种关注边界的水下图像语义分割方法及真实水下场景语义分割数据集

马志伟[1]　李豪杰[1]　王智慧[1]　于丹[1]　王天一[1]　顾滢双[1]　樊鑫[1]　罗钟铉[1]

[1](大连理工大学软件学院 大连 中国 116621)

**摘　要**　随着水下生物抓取技术的不断发展，高精度的水下物体识别与分割成为了挑战。目前已有的水下目标检测技术仅能给出物体的大体位置，无法提供物体轮廓等更加细致的信息，严重影响了抓取效率。为了解决这一问题，本文首先标注并建立了第一个真实场景水下语义分割数据集（DUT-USEG：DUT Underwater Segmentation Dataset）。DUT-USEG 数据集包含 6617 张图像，其中 1487 张具有语义分割和实例分割标注，剩余 5130 张图像具有目标检测框标注。基于此数据集，我们提出了一个关注边界的半监督水下语义分割网络（US-Net: Underwater Segmentation Network）。该网络通过设计伪标签生成器和边界检测子网络，实现了对水下物体与背景之间边界的精细学习，提升了边界区域的分割效果。实验表明，我们提出的方法在 DUT-USEG 数据集的海参，海胆，海星三个类别上提升了 6.7%，达到了目前最好的分割精度。DUT-USEG 数据集将会发布在 https://github.com/baxiyi/DUT-USEG。

**关键词**　水下生物抓取；水下语义分割；半监督学习；弱监督学习；边界检测

**中图法分类号**　****　　DOI 号 *投稿时不提供 DOI 号*

## An Underwater Image Semantic Segmentation Method Focusing on Boundaries and a Real Underwater Scene Semantic Segmentation Dataset

MA Zhiwei[1]　LI Haojie[1]　WANG Zhihui[1]　YU Dan[1]　WANG Tianyi[1]　GU Yingshuang[1]　FAN Xin[1]　LUO Zhongxuan[1]

[1](Department of software engineering, Dalian University of Technology, Dalian 116621, China)

**Abstract**　With the development of underwater object grabbing technology, underwater object recognition and segmentation of high accuracy has become a challenge. The existing underwater object detection technology can only give the general position of an object, unable to give more detailed information such as the outline of the object, which seriously affects the grabbing efficiency. To address this problem, we label and establish the first underwater semantic segmentation dataset of real scene(DUT-USEG：DUT Underwater Segmentation Dataset). The DUT-USEG dataset includes 6617 images, 1487 of which have semantic segmentation and instance segmentation annotations, and the remaining 5130 images have object detection box annotations. Based on this dataset, we propose a semi-supervised underwater semantic segmentation network focusing on the boundaries(US-Net: Underwater Segmentation Network). By designing a pseudo label generator and a boundary detection subnetwork, this network realizes the fine learning of boundaries between underwater objects and background, and improves

---

the segmentation effect of boundary areas. Experiments show that the proposed method improves by 6.7% in three categories of holothurian, echinus, starfish in DUT-USEG dataset, and achieves state-of-the-art results. The DUT-USEG dataset will be released at https://github.com/baxiyi/DUT-USEG.
**Key words** underwater object grabbing；underwater semantic segmentation；semi-supervised learning；weakly supervised learning；boundary detector

# 1 引言

近年来，随着水下机器人目标抓捕需求的不断上升，有诸多研究开始关注水下机器视觉领域。目前的水下视觉研究主要包括水下图像增强[1-3]和水下目标检测[4-6]，水下图像增强的目的是提高水下图像的质量，水下目标检测的目的是为水下机器人的抓捕提供物体的识别与定位。然而目标检测只能够为物体提供一个矩形包围框，而无法给出物体轮廓等更加细致的信息，尤其是当水下环境和物体本身难以区分时，即使有目标检测的结果仍然很难完成水下目标的精准抓取，而计算机视觉中的语义分割任务则能够很好的完成物体与其背景之间的区分。因此本文专注于水下机器视觉的一个新研究问题——水下图像语义分割。

为了进行该项研究，我们提出了第一个真实场景水下语义分割数据集（DUT-USEG），该数据集包括6617张水下图像，包含了海参，海胆，扇贝，海星四个类别，其中1487张图像具有我们手工添加的语义分割标注和实例分割标注，剩余的5130张图像具有目标检测框标注（这些目标检测标注是Liu在[7]中完成的）。

基于该数据集，我们进一步进行了水下图像语义分割的研究，提出了一个包括伪标签生成器和边界检测子网络的半监督语义分割网络（US-Net）。针对水下图像中物体与水下环境难以区分的问题，我们设计了一个边界检测网络，该网络通过融合多个不同尺度的特征图来完成类边界的学习。此外，为了有效利用数据集中无语义分割标注的数据，我们设计了一个伪标签生成器，通过已有的分类标注和框标注分别得到类激活图和框注意力图，将二者融合后通过阈值筛选得到语义分割的伪标签，利用生成的伪标签和已有的标注数据共同监督边界检测网络的学习，伪标签的生成一定程度上解决了监督信息稀少的问题。实验表明，本文提出的方法在DUT-USEG数据集上的精度上优于目前已有的其它方法。

本文的贡献主要如下：

（1）提出了一个真实场景水下语义分割数据集（DUT-USEG），具我们所知，这是第一个开源的真实场景水下图像语义分割数据集。

（2）针对水下语义分割问题，提出了一个半监督的语义分割网络（US-Net），该方法通过设计一个伪标签生成器和边界检测子网络以解决水下物体与背景之间边界难以区分的问题。

（3）实验表明，本文提出的US-Net网络在DUT-USEG数据集上达到了目前最好的语义分割结果。

# 2 相关工作

## 2.1 水下目标检测

水下目标的抓捕首先需要对水下生物进行定位，因此目前有很多工作对水下目标检测进行了研究。Li等人[5]使用Fast RCNN网络进行鱼类的检测和识别。Villon[6]等人将深度学习方法与定向网格直方图（HOG）+支持向量机（SVM）方法在珊瑚礁鱼类检测中进行了比较，证明了深度学习方法在水下目标检测中的优越性。在最近的研究中，Chen[4]等人提出了一个利用多个高级特征图以改进小物体目标检测的网络，并设计了样本加权损失函数来监督网络的学习。也有一些工作[7,8]针对水下目标检测数据缺少的问题，提出了相关的数据集。与我们工作最相关的是Liu[7]最近提出的数据集，我们的数据集是在该数据集的基础上制作的。

与这些水下目标检测工作不同，我们专注于水下图像的语义分割，我们认为语义分割得到的像素级分割结果对于完成更精确的水下目标抓取能够提供帮助。

## 2.2 基于框的弱监督语义分割

基于全卷积神经网络（FCN）[9]的语义分割方法已经发展的很成熟了。但是由于语义分割的标注成本较高，目前有很多工作开始研究弱监督语义分割，与本文较为相关的方法是基于框进行弱监督语义分割。He等人[10]提出了一种以框为监督，在自动

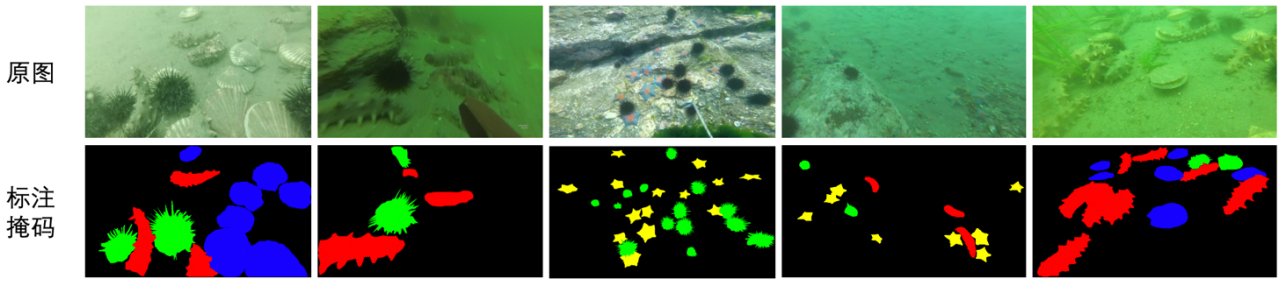

图 1 语义分割的标注样例。第一行为原图，第二行为原图对应的标注掩码。

生成区域建议和训练卷积网络之间进行迭代的方式来逐步改善分割掩码的方法。Khoreva[11]等人指出，通过 GrabCut[12]或 MCG[13]等传统方法首先得到伪标签，经过单次训练就可以得到很好的效果。在最近的研究中，Song 等人[14]和 Li 等人[15]通过不同的方式对伪标签质量进行改善，从而达到了更好的分割结果。Zhang 等人[16]的研究使用高斯注意力图和引导梯度反向传播图作为额外输入，为准确分割提供先验线索。

与以上方法不同，本文的方法主要关注于水下图像的语义分割。由于水下图像环境的复杂性以及水下能见度较低等问题，很多水下物体与周围的环境之间的区分度很低，导致物体的边界很难与背景区分开。同时，由于水下物体之间经常出现彼此挨得很近的情况，物体与物体之间的边界也很难分割，图 2 展示了这两个问题的样例图片。因此，将现有的一些方法直接应用于水下图像的语义分割会出现很多问题。GrabCut 等传统算法容易将框内的整个区域都当做前景区域，而基于深度学习的方法则经常分割不出目标物体，或者只分割出目标物体的一小部分，如图 6 所示。为了解决之前方法存在的问题，我们的方法设计了一个边界检测子网络来检测类之间的边界，从而达到更好区分前景与背景的目的。

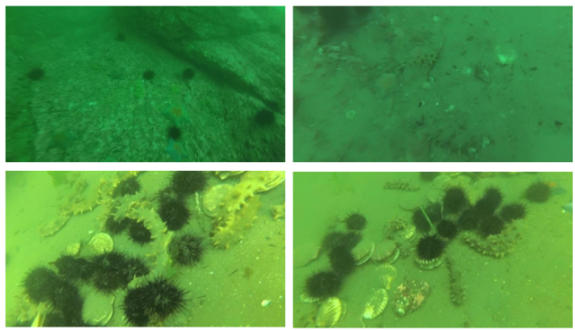

图 2 边界难以分割的样例图片。第一行的两张图展示了物体与背景之间的边界难以区分的问题；第二行展示了物体之间边界难以区分的问题。

## 3 数据集

### 3.1 数据集收集与标注

由于目前还没有开源的水下图像语义分割数据集，为了完成水下语义分割的研究，我们以之前的目标检测水下数据集为基础构建了一个水下图像分割数据集。

我们的数据来源于 Liu 等人[7]提出的水下目标检测数据集。根据图像拍摄的清晰度，我们从该数据集的 7782 张图像中挑选了 6617 张作为我们的数据集，并对其中的 1487 张进行了语义分割和实例分割的标注，剩余的 5130 张保留了原数据集的目标检测标注。

我们的标注包括海参，海胆，扇贝，海星四个类别，使用的是 COCO 数据集[17]的标注格式，为了方便语义分割任务的研究，我们像 PASCAL VOC 2012[18]一样提供了语义分割的掩码图片。图 1 展示了语义分割标注的一些样例，可以看到我们的数据集包含了多种不同的水下场景。

### 3.2 数据集的相关统计

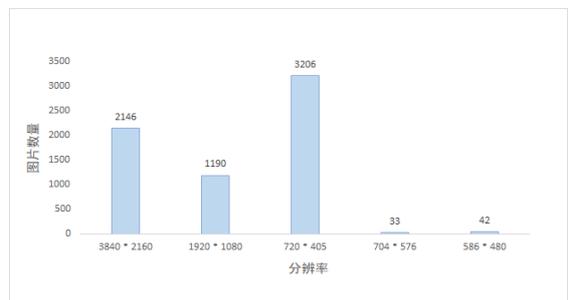

图 3 数据集中不同分辨率图像的统计情况

由于采集设备的不同，我们的数据集中包含了多种不同分辨率的图像，最低分辨率为 586*480，最高的分辨率达到了 3840*2160。这种分辨率的差异也是水下目标实际抓捕过程中经常会遇到的问

题。图 3 展示了我们的数据集中不同分辨率图像的数量统计情况。可以看到 1K 以上的高分辨率的图像占到了图像总数的一半左右，由于在训练过程中进行下采样造成的细节信息损失问题，高分辨率图像的分割也是语义分割问题中的一个难题[19]。

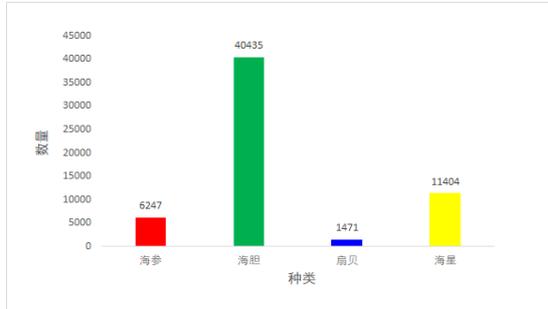

图 4 数据集中不同类别对应的实例数量

我们的数据集中包含了 4 类水下生物，分别是海参，海胆，扇贝和海星，图 4 展示了四个类别对应实例所占的数量。可以看到我们的数据集存在一定的类别不均衡问题，海胆的实例数目很多而扇贝的实例数目较少。

# 4 方法

本文提出的 US-Net 整体网络框架如图 5 所示。它包含一个伪标签生成器和一个边界检测网络，我们将分别在 4.1 和 4.2 小节进行详细介绍。

## 4.1 伪标签生成器

由于在训练过程中仅有部分数据为语义分割标注数据，因此对于没有语义分割标注的数据，我们通过分类和目标检测框的标注信息为它们生成伪标签。伪标签的生成流程如图 5(a)所示。

### 4.1.1 类激活图

为了结合分类网络的信息，我们使用一种类激活图的方法[20]，该方法首先将分类网络最后一层卷积输出的特征图进行全局平均池化，再通过一个全连接层得到各个类别对原特征图的权重，使用该权重对原特征图进行加权求和从而得到原特征图中的各个类的判别性区域。

假设分类网络的最后一层卷积得到的特征图维度为$K$，令$f_k(x,y)$表示在特征图第$k$层的$(x,y)$位置的激活值。将该特征图进行全局平均池化后得到$F_k = \sum_{(x,y)} f(x,y)$。假设在全连接层中$F_k$对应的权重为$w_k^c$，$M_c$表示在类别$c$上的类激活图，则：

$$M_c(x,y) = \sum_k w_k^c f_k(x,y) \quad (1)$$

该方法得到的效果如图 5 中的类激活图所示。

### 4.1.2 基于框的高斯注意力图

对于框标注信息的利用，我们借鉴了 Zhang 等人[16]的做法，使用一个框中心为均值，框的长和宽为方差的二维高斯分布作为高斯注意力图来表示框包含的位置信息。

具体来说，对于一个框$B^i:[x^i,y^i,w^i,h^i]$，其中$(x^i,y^i)$为框中心的坐标，$w^i,h^i$为框的宽和高，那么框$B^i$的高斯注意力图可以表示为：

$$G(x,y) = \frac{1}{2\pi\sigma_1\sigma_2\sqrt{1-\rho^2}} e^{-\frac{1}{2(1-\rho^2)}F(x,y)} \quad (2)$$

$$F(x,y) = \frac{(x-\mu_1)^2}{\sigma_1^2} - 2\rho\frac{(x-\mu_1)(y-\mu_2)}{2\sigma_1\sigma_2} + \frac{(y-\mu_2)^2}{\sigma_2^2} \quad (3)$$

其中$\mu_1 = x^i, \mu_2 = y^i, \sigma_1 = w^i, \sigma_2 = h^i$，$\rho$为相关系数。

### 4.1.3 伪标签生成

为了得到更加可靠的伪标签，我们首先对类激活图和框注意力图进行融合，在融合过程中，考虑到分类网络的性能上限问题，会有一些图片没有被正确分类，对于分类错误的情况，我们直接使用该图的真实类别对应的框生成高斯注意力图，用来替代在该类别上的激活图。对于正确的分类，仍然需要考虑两种情况，第一种情况是类别对应的框内存在类激活区域（我们使用区域内的激活值均大于某个阈值作为判断），这种情况下我们直接使用类激活图和框注意力图的点对点乘积作为结果，这样可以让类激活图中超出物体本身的激活区域（在框的边缘及超出框的部分）得到有效抑制，第二种情况是当框内不存在类激活区域时，我们仍然直接使用框注意力图作为结果。

具体来说，对于一张图像$I$及其对应的一个框$B^j:[x^j,y^j,w^j,h^j]$，$(x^j,y^j)$为框左上角的坐标，$w^j,h^j$为框的宽和高。假设$B^j$对应的真实类别为$c_g$，$I$在分类网络中的预测类别集合为 C，则融合结果$S_{c_g}^j$的计算公式为：

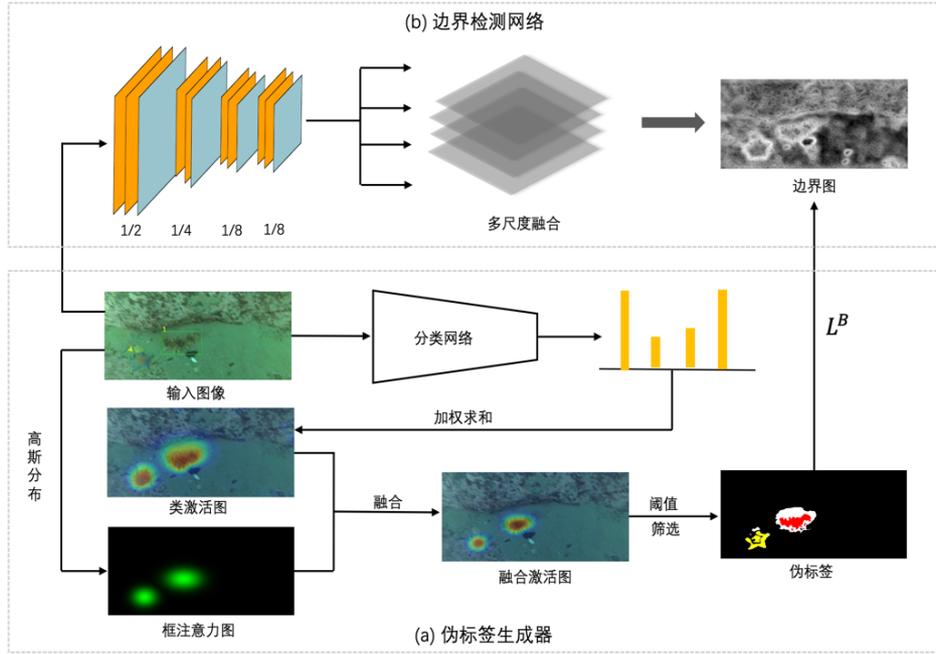

图 5  US-Net 网络框架，分为伪标签生成器和边界检测网络。(a)为伪标签生成器，通过融合类激活图和框注意力图生成伪标签，并通过$L^B$损失函数来监督边界网络的学习。(b)为边界检测网络，融合了多尺度特征来更好的学习边界图。

$$S_{c_g}^j(x,y) = \begin{cases} G^j(x,y), & if\ c_g \notin C \\ G^j(x,y), & if\ c_g \in C\ and\ cond_1 \\ G^j(x,y) * M_c(x,y), & if\ c_g \in C\ and\ cond_2 \end{cases} \quad (4)$$

$$cond_1 = \{\forall x,y, M_c(x,y) < \varepsilon\}_1 \quad (5)$$

$$cond_2 = \{\exists x,y, M_c(x,y) \geq \varepsilon\}_1 \quad (6)$$

其中 $x \in (x^j, x^j+w^j), y \in (y^j, y^j+h^j)$，$\{cond_i\}_1$ 表示一个布尔值，满足{}中的条件则为$true$，否则为$false$，$G^j$和$M_c$为公式(1)和公式(2)中得到的结果，$\varepsilon$为设定的阈值。

在完成融合之后，我们根据前背景阈值对(4)中生成的融合激活图$S$进行筛选从而得到伪标签。依据实际经验，我们使用 0.3 作为前景阈值，0.05 作为背景阈值。也就是说当某点类的激活值大于 0.3 时，我们认为该点属于这个类，如果有多个点均大于 0.3，则取激活值最大的类；如果某点各个类激活值均小于 0.05，则认为其为背景；我们忽略不满足以上两个条件的其它像素，因为我们认为这些像素是不可靠的，无法作为监督训练的标签。

### 4.2 边界检测网络

为了解决水下图像的边界分割问题，我们受[21]中的方法的启发设计了一个边界检测网络，网络的结构如图 5(b)所示。该网络结构通过融合不同尺度的信息来更好的学习边界图。

对于有语义分割掩码标注的数据，我们直接使用其标注作为标签。对于只有框标注的数据，我们利用伪标签生成器得到它们的伪标签。根据获得的掩码标签中像素间类别的关系，我们将像素划分为同类像素对和不同类像素对两个集合。为了减小不必要的计算，我们只对距离较近的像素进行类关系的判断，使用一个距离阈值进行限制。

具体来说，对于两个像素点$p_i$和$p_j$：

$$P = \{(i,j) |\ ||x_i - x_j|| < \gamma, \forall i \neq j\} \quad (7)$$

$$P^+ = \{(i,j) | L(p_i) = L(p_j), (i,j) \in P\} \quad (8)$$

$$P^- = \{(i,j) | L(p_i) \neq L(p_j), (i,j) \in P\} \quad (9)$$

其中$\gamma$为两个像素之间的最大欧式距离，$L$为之前得到的伪标签。这样得到的$P^+$即为属于同类的像素对（包括同属于前景类或者同属于背景类），$P^-$为不属于同一类的像素对。

有了以上集合划分后，我们根据这些像素对来设计边界检测器网络的损失函数。如果两个像素属于同类像素，那么在它们之间的连线上应该均为该类别的像素，即它们之间不应该存在边界，反之，应该存在边界，因此我们使用以下公式来衡量两个

像素$p_i$和$p_j$之间的语义相关性$a_{ij}$：

$$a_{ij} = 1 - \max B(x_k), k \in l(i,j) \quad (10)$$

其中$l(i,j)$表示$p_i$和$p_j$连线上的像素点，$B(x_k)$为$x_k$点属于边界的概率值，也就是说$B$是我们最终要得到的边界概率图。当$p_i$和$p_j$为同类像素，我们希望$a_{ij}$接近1，反之，希望$a_{ij}$接近0，根据这个关系，我们将最终的损失函数设计成一个二元的交叉熵损失函数。

$$L^B = -\left(\sum_{(i,j) \in P^+} \frac{\log a_{ij}}{|P^+|} + \sum_{(i,j) \in P^-} \frac{\log(1-a_{ij})}{|P^-|}\right) \quad (11)$$

在得到边界概率图后，我们使用[21]中随机游走的方式逐步迭代来完善公式(4)中得到的融合激活图，使得类的激活区域向边界进行扩张，从而得到分割掩码，再经过条件随机场[22]后处理得到最终的分割结果。

# 5 实验结果

为了验证我们的方法在真实水下场景 DUT-USEG 数据集上的有效性，我们将有语义分割标注的数据一部分划分为测试集，并将另一部分和只有框标注的数据作为训练集，具体来说，我们的训练集包含 5863 张图像，其中 733 张为有语义分割标注的图像，约占训练集总数的 1/8，测试集包含 754 张图像。

## 5.1 实现细节

### 5.1.1 超参数设置

在融合类激活图和框注意力图的时候，我们需要设定阈值来判断框内是否有激活区域，即公式(5)和公式(6)中的$\varepsilon$，本文实验中我们取$\varepsilon = 0.3$。

在生成伪标签时，我们需要设置前景阈值和背景阈值，在本文实验中我们分别将前景阈值和背景阈值分别设置为 0.3 和 0.05。

在实现边界检测网络的损失函数时，需要设置像素点对之间的最大距离，即公式（7）中的$\gamma$，本文实验中将$\gamma$设置为5。

### 5.1.2 训练细节

本文的所有训练过程是在一块 Tesla V100-SXM2-32GB 显卡上完成的。对于获取类激活图的分类网络，我们将初始学习率设置为 0.01，并使用学习率衰减的策略[23]训练了 30 个 epoch；对于边界检测网络，我们将初始学习率设置为 0.1，同样使用学习率衰减策略训练 30 个 epoch，训练中使用的优化方法是随机梯度下降法。

### 5.1.3 其它细节

在生成类激活图的时候，为了得到更好的效果，我们将原图片进行了不同尺度的缩放得到多张类激活图，并进行加和平均。

在测试过程中，我们观察到由于测试数据并不在分类网络的训练数据中，因此在测试数据上得到的类激活图效果很差，所以测试过程中我们舍弃类激活图，直接使用框的高斯注意力图作为激活图并使用边界图进行完善。

## 5.2 对比实验

我们将我们的方法（US-Net）和之前的同类方法（测试阶段具有框的先验知识）进行了对比，对比的方法包括 GrabCut[12]和 Zhang 等人[16]的方法（GGANet）。

由于 GrabCut 在无法分割出前景时会将整个框内的所有像素作为前景，为了对比的公平性，我们将 GGANet 和我们方法的预测结果也做了同样的处理。对比实验的结果如表 1 所示，我们使用的是语义分割中最常用的评价指标 mIOU（平均交并比）。由于 GGANet 的网络在我们的数据集上训练后无法收敛，对比实验中使用的是原文中在 Pascal VOC 2012 上训练得到的用于分割未见类的模型。

可以看到我们的方法在海参，海胆，海星三个类别上都达到了最好，最终的平均值也达到了最好，但在扇贝类别上低于 GrabCut，可能原因在于我们的数据集中扇贝的数量较少(如图 4 所示)，而 GrabCut 是一种传统算法，性能不会受到训练数据量的影响。图 6 展示了我们的方法和其它两种方法的可视化结果，可以看到由于边界检测器的作用，我们的方法在边界区域有更好的分割效果。不同于 Grabcut 经常将框内大部分区域分割为前景，我们的分割结果更加接近物体的实际形状，这使得我们的方法对水下生物的抓取具有更强的指导意义。

此外本文的实验是基于目标检测标注框进行的，测试阶段同样也可以在已有目标检测网络结果的基础上进行，从而使得我们的方法在实际水下目标抓取中具有更灵活的应用价值。

## 5.3 消融实验

为了验证边界检测网络对实验结果的影响，我

表 1 对比实验结果

| | 海参 | 海胆 | 扇贝 | 海星 | 平均值（含背景） |
|---|---|---|---|---|---|
| GrabCut | 52.5 | 63.8 | **77.8** | 58.2 | 70.1 |
| GGANet | 32.1 | 42.9 | 52.8 | 34.5 | 52.1 |
| US-Net | **54.9** | **67.7** | 70.4 | **63.6** | **71.1** |

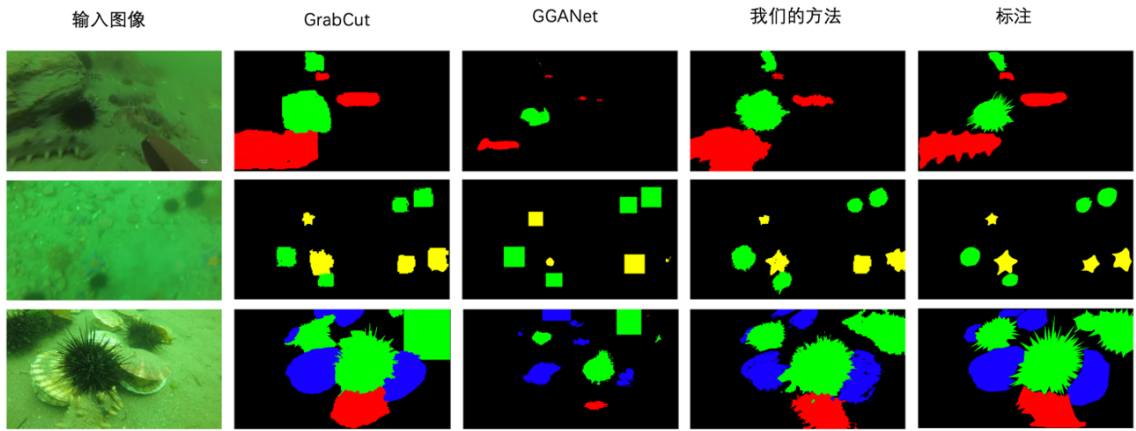

图 6 可视化结果

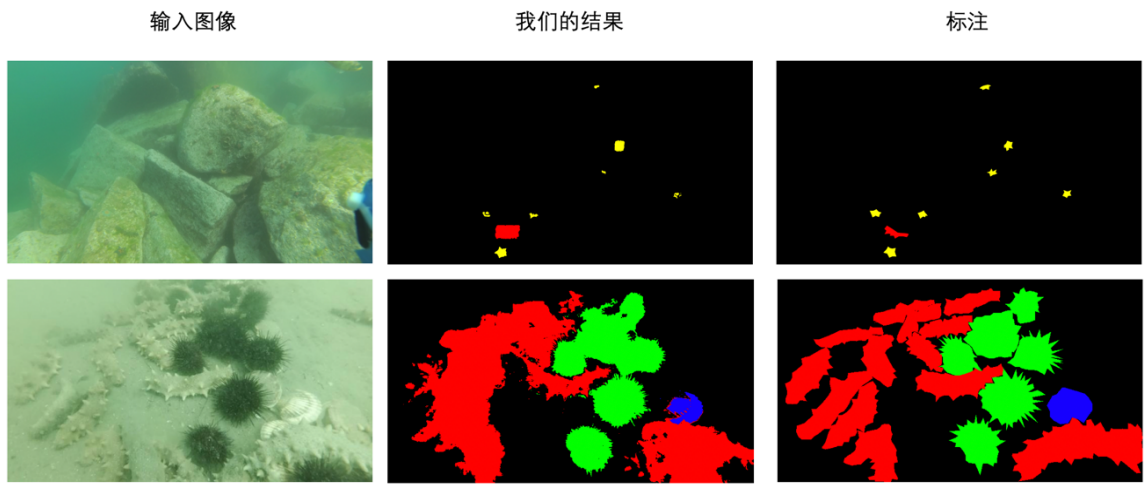

图 7 失败案例

们进行了边界检测器网络的消融实验，如表 2 所示。可以看到边界检测网络对实验结果有明显提升。

表 2 消融实验结果

| 方法 | mIOU |
|---|---|
| 只用高斯注意力图 | 68.8 |
| 高斯注意力图+边界检测网络（我们的方法） | 71.1 |

## 5.4 局限性分析

由于在我们的数据集中存在大量的高分辨率图像，见图 3。在训练和测试过程中需要对这些图像进行下采样，以减小计算量，这就导致一些小物体在下采样过程中损失了较多信息，所以我们的方法在部分高分辨率图像的小物体上表现较差，如图 7 第一行图片所示。我们准备在之后的工作中尝试更好的处理高分辨率图像。

此外，由于我们的边界检测网络检测的只是类别之间的边界，导致一些同类但相距很近的物体之间的边界难以被区分，如图 7 第二行图片所示。未来我们会尝试设计一个检测实例之间边界的网络来解决这一问题。

# 6 结论

为了推进水下图像语义分割方向的研究，本文提出了一个水下语义分割数据集(DUT-USEG)，该数据集包含6617张水下图像，其中1487张具有语义分割和实例分割标注。同时，本文提出了一个半监督语义分割网络(US-Net)，主要包含伪标签生成器和边界检测子网络，用来解决水下图像中物体与背景之间边界难以区分的问题。实验表明该方法在目前的同类方法中达到了最高精度。

未来，我们会在水下图像分割领域做更深入的研究，我们将力争解决目前方法中存在的高分辨率小物体分割不准确和同类物体边界分割不准确的问题。

## 参 考 文 献

[1] Li C, Guo C, Ren W, et al. An underwater image enhancement benchmark dataset and beyond[J]. IEEE Transactions on Image Processing, 2019, 29: 4376-4389

[2] Li C, Anwar S, Porikli F. Underwater scene prior inspired deep underwater image and video enhancement[J]. Pattern Recognition, 2020, 98: 107038.

[3] Guo Y, Li H, Zhuang P. Underwater image enhancement using a multiscale dense generative adversarial network[J]. IEEE Journal of Oceanic Engineering, 2019, 45(3): 862-870.

[4] Chen L, Liu Z, Tong L, et al. Underwater object detection using Invert Multi-Class Adaboost with deep learning[C]//2020 International Joint Conference on Neural Networks (IJCNN). IEEE, 2020: 1-8.

[5] Li X, Shang M, Qin H, et al. Fast accurate fish detection and recognition of underwater images with fast r-cnn[C]//OCEANS 2015-MTS/IEEE Washington. IEEE, 2015: 1-5.

[6] Villon S, Chaumont M, Subsol G, et al. Coral reef fish detection and recognition in underwater videos by supervised machine learning: Comparison between Deep Learning and HOG+ SVM methods[C]//International Conference on Advanced Concepts for Intelligent Vision Systems. Springer, Cham, 2016: 160-171.

[7] Liu, Chongwei, Haojie Li, Shuchang Wang, et al. A Dataset And Benchmark Of Underwater Object Detection For Robot Picking[J]. arXiv preprint arXiv: 2106.05681(2021)

[8] Jian M, Qi Q, Dong J, et al. The OUC-vision large-scale underwater image database[C]//2017 IEEE International Conference on Multimedia and Expo (ICME). IEEE, 2017: 1297-1302.

[9] Long J, Shelhamer E, Darrell T. Fully convolutional networks for semantic segmentation[C]//Proceedings of the IEEE conference on computer vision and pattern recognition. 2015: 3431-3440.

[10] Dai J, He K, Sun J. Boxsup: Exploiting bounding boxes to supervise convolutional networks for semantic segmentation[C]//Proceedings of the IEEE international conference on computer vision. 2015: 1635-1643.

[11] Khoreva A, Benenson R, Hosang J, et al. Simple does it: Weakly supervised instance and semantic segmentation[C]//Proceedings of the IEEE conference on computer vision and pattern recognition. 2017: 876-885.

[12] Rother C, Kolmogorov V, Blake A. " GrabCut" interactive foreground extraction using iterated graph cuts[J]. ACM transactions on graphics (TOG), 2004, 23(3): 309-314.

[13] Pont-Tuset J, Arbelaez P, Barron J T, et al. Multiscale combinatorial grouping for image segmentation and object proposal generation[J]. IEEE transactions on pattern analysis and machine intelligence, 2016, 39(1): 128-140.

[14] Song C, Huang Y, Ouyang W, et al. Box-driven class-wise region masking and filling rate guided loss for weakly supervised semantic segmentation[C]//Proceedings of the IEEE/CVF Conference on Computer Vision and Pattern Recognition. 2019: 3136-3145.

[15] Lee J, Yi J, Shin C, et al. BBAM: Bounding Box Attribution Map for Weakly Supervised Semantic and Instance Segmentation[J]. arXiv preprint arXiv:2103.08907, 2021.

[16] Zhang P, Wang Z, Ma X, et al. Learning to Segment Unseen Category Objects using Gradient Gaussian Attention[C]//2019 IEEE International Conference on Multimedia and Expo (ICME). IEEE, 2019: 1636-1641.

[17] Lin T Y, Maire M, Belongie S, et al. Microsoft coco: Common objects in context[C]//European conference on computer vision. Springer, Cham, 2014: 740-755.

[18] Everingham M, Van Gool L, Williams C K I, et al. The pascal visual object classes (voc) challenge[J]. International journal of computer.

[19] Cheng H K, Chung J, Tai Y W, et al. CascadePSP: Toward Class-Agnostic and Very High-Resolution Segmentation via Global and Local Refinement[C]//Proceedings of the IEEE/CVF Conference on Computer Vision and Pattern Recognition. 2020: 8890-8899.

[20] Zhou B, Khosla A, Lapedriza A, et al. Learning deep features for discriminative localization[C]//Proceedings of the IEEE conference on computer vision and pattern recognition. 2016: 2921-2929.

[21] Ahn J, Cho S, Kwak S. Weakly supervised learning of instance segmentation with inter-pixel relations[C]//Proceedings of the IEEE/CVF Conference on Computer Vision and Pattern Recognition. 2019: 2209-2218.

[22] Krähenbühl P, Koltun V. Efficient inference in fully connected crfs with gaussian edge potentials[J]. Advances in neural information processing systems, 2011, 24: 109-117.

[23] Liu W, Rabinovich A, Berg A C. Parsenet: Looking wider to see better[J]. arXiv preprint arXiv:1506.04579, 2015.